\documentclass{article}

\usepackage{PRIMEarxiv}

\usepackage[utf8]{inputenc} 
\usepackage[T1]{fontenc}    
\usepackage{hyperref}       
\usepackage{url}            
\usepackage{booktabs}       
\usepackage{amsfonts}       
\usepackage{nicefrac}       
\usepackage{microtype}      
\usepackage{lipsum}
\usepackage{fancyhdr}       
\usepackage{graphicx}       
\graphicspath{{media/}}     

\usepackage{tikz}
\usepackage{comment}
\usepackage{amsmath,amssymb} 
\usepackage{color}
\usepackage[accsupp]{axessibility}  

\title{Neural Density-Distance Fields
}

\author{
  Itsuki Ueda \\
  University of Tsukuba \\
  \texttt{ueda.itsuki@image.iit.tsukuba.ac.jp} \\
   \And
  Yoshihiro Fukuhara \\
  Waseda University \\
  \texttt{f\_yoshi@ruri.waseda.jp} \\
   \And
  Hirokatsu Kataoka \\
  National Institute of Advanced Industrial Science and Technology \\
  \texttt{hirokatsu.kataoka@aist.go.jp} \\
   \And
  Hiroaki Aizawa \\
  Hiroshima University \\
  \texttt{hiroaki-aizawa@hiroshima-u.ac.jp} \\
   \And
  Hidehiko Shishido \\
  University of Tsukuba \\
  \texttt{shishido.hidehiko@image.iit.tsukuba.ac.jp} \\
   \And
  Itaru Kitahara \\
  University of Tsukuba \\
  \texttt{kitahara.itaru@image.iit.tsukuba.ac.jp} \\
}

\begin{document}
\maketitle

\begin{abstract}
The success of neural fields for 3D vision tasks is now indisputable. Following this trend, several methods aiming for visual localization (e.g., SLAM) have been proposed to estimate distance or density fields using neural fields. However, it is difficult to achieve high localization performance by only density fields-based methods such as Neural Radiance Field (NeRF) since they do not provide density gradient in most empty regions. On the other hand, distance field-based methods such as Neural Implicit Surface (NeuS) have limitations in objects' surface shapes. This paper proposes Neural Density-Distance Field (NeDDF), a novel 3D representation that reciprocally constrains the distance and density fields. We extend distance field formulation to shapes with no explicit boundary surface, such as fur or smoke, which enable explicit conversion from distance field to density field. Consistent distance and density fields realized by explicit conversion enable both robustness to initial values and high-quality registration. Furthermore, the consistency between fields allows fast convergence from sparse point clouds. Experiments show that NeDDF can achieve high localization performance while providing comparable results to NeRF on novel view synthesis. 
The code is available at \url{https://github.com/ueda0319/neddf}.
\end{abstract}

\section{Introduction}

\begin{figure}[!hbt]
    \begin{center}
    \includegraphics[width=0.9\linewidth]{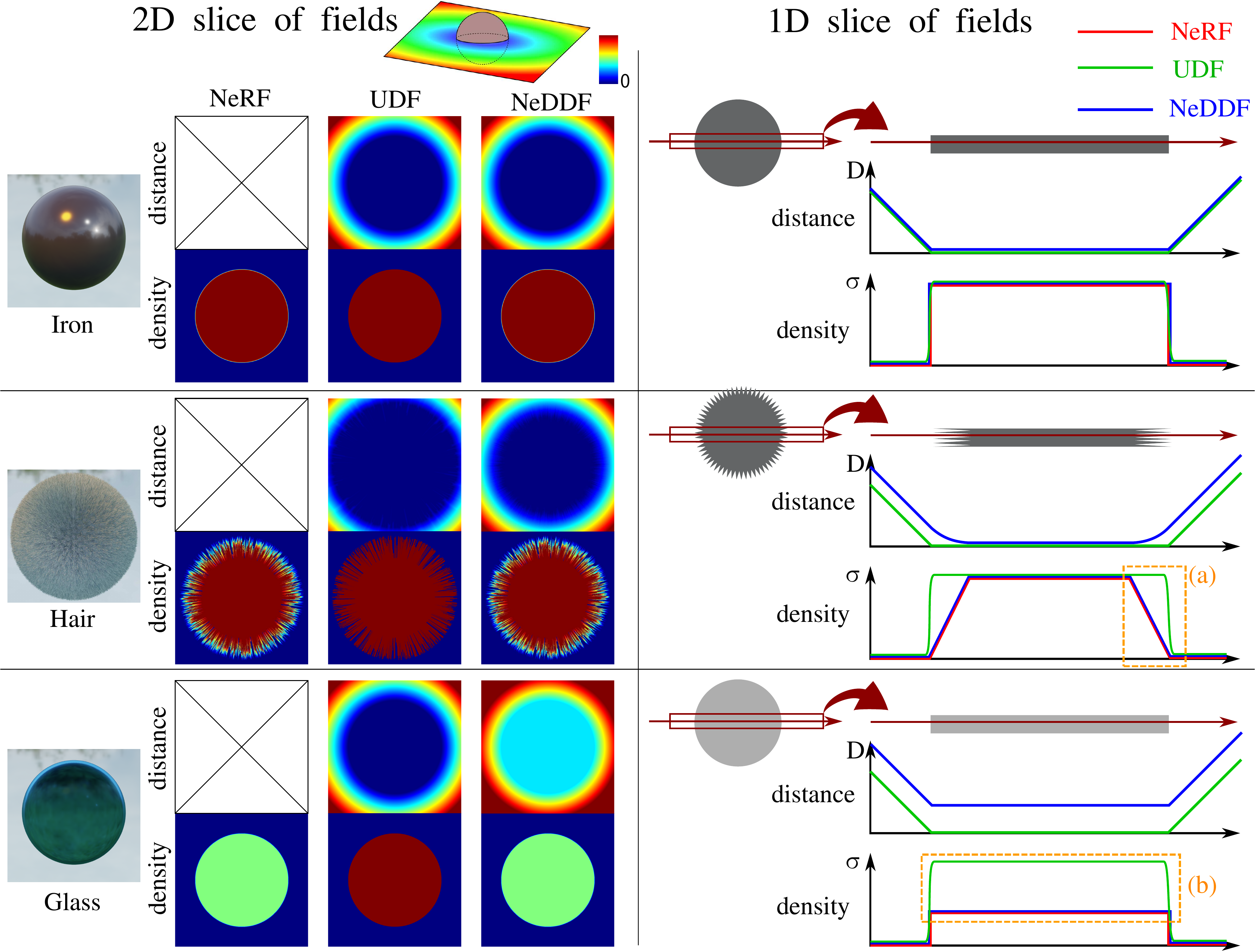}
    \end{center}
    \caption{(left) Visualization of the 2D slice for each field with iron, hair, and glass spheres as examples. (right) Plots of 1D slices for each field. NeRF provides no distance information. Unsigned Distance Field (UDF) cannot handle some cases correctly, such as (a) ambiguous density changes such as a hairball or (b) low densities such as a glass ball. Proposed NeDDF can represent both cases properly.}
    \label{fig:FieldReps}
\end{figure}

Representing 3D shapes using coordinate-based neural networks, also known as neural fields~\cite{xie2021neuralfield} have recently attracted attention as an alternative to using point clouds, voxels, meshes, and others~\cite{NeRF,Nerfies,DNeRF,deepsdf,UNISURF,UDF,NeuS}. Neural Radiance Fields (NeRF)~\cite{NeRF}, in particular, have shown impressive quality for tasks such as novel view-synthesis. However, since NeRF has limited regions with smooth spatial density and color, many conventional methods still require good initial values for registration and localization tasks. This paper proposes a distance field representation that is reciprocally constrained to the density field, named Neural Density-Distance Field (NeDDF). NeDDF achieves robust localization with distance fields while providing object reconstruction quality comparable to NeRF.

\begin{figure}[!ht]
    \begin{center}
    \includegraphics[width=0.5\linewidth]{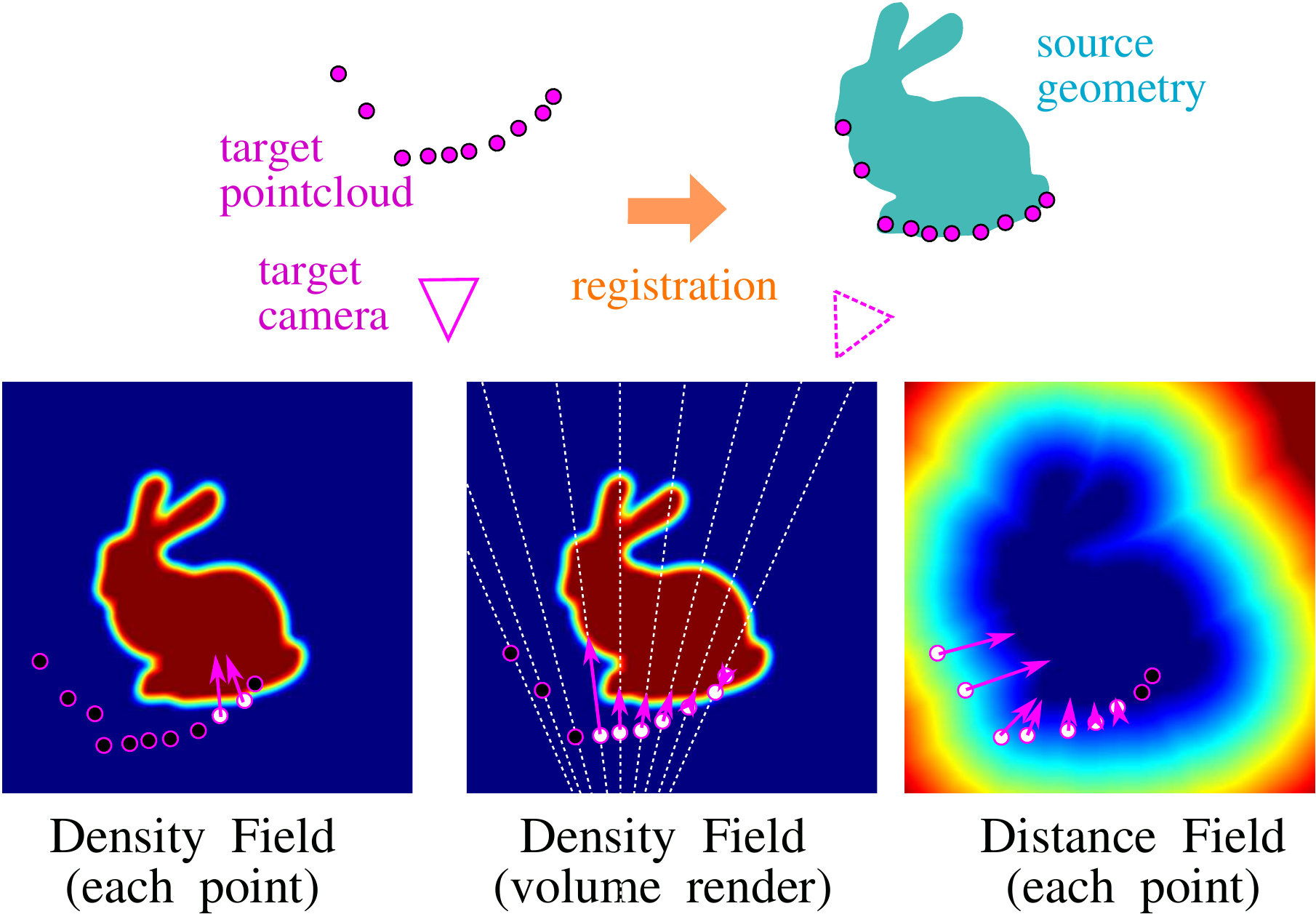}
    \end{center}
    \caption{Registration following in each representation: white and black points (small circles) denote the points where the gradient direction is available or not, respectively. In the density field (left), point-by-point gradients cannot be obtained in most regions except for the boundary. In volume rendering of the density field (center), it is impossible to obtain the gradient component in the vertical direction of view. In the distance field (right), we can obtain both the gradient direction and the residuals for each point.}
    \label{fig:trackable}
\end{figure}

As shown in Fig. \ref{fig:FieldReps}, there are two main types of 3D shape representation in neural fields: density field used in NeRF~\cite{NeRF} and distance field used in NeuS~\cite{NeuS}. 
Density field has high expressiveness for translucent objects, such as smoke and water, and high-frequency shapes, such as hair.
However, in most areas except the boundary, the gradient of the field is zero.
This makes it difficult to set up a convex objective function in a problem setting such as registration, as shown in Fig.~\ref{fig:trackable}.
Distance field provides the gradient over a wide range even after the optimization converges.
Thus, we can establish objective functions with high convexity in registration.
The field can be learned from the image via volume rendering by defining a conversion equation from distance to density.
For example, NeuS assumes that the density follows a logistic distribution close to the object surface.
On the other hand, since we assume explicit boundaries, the convertible density field is tightly restricted.

\begin{figure}[ht]
    \begin{center}
    \includegraphics[width=0.7\linewidth]{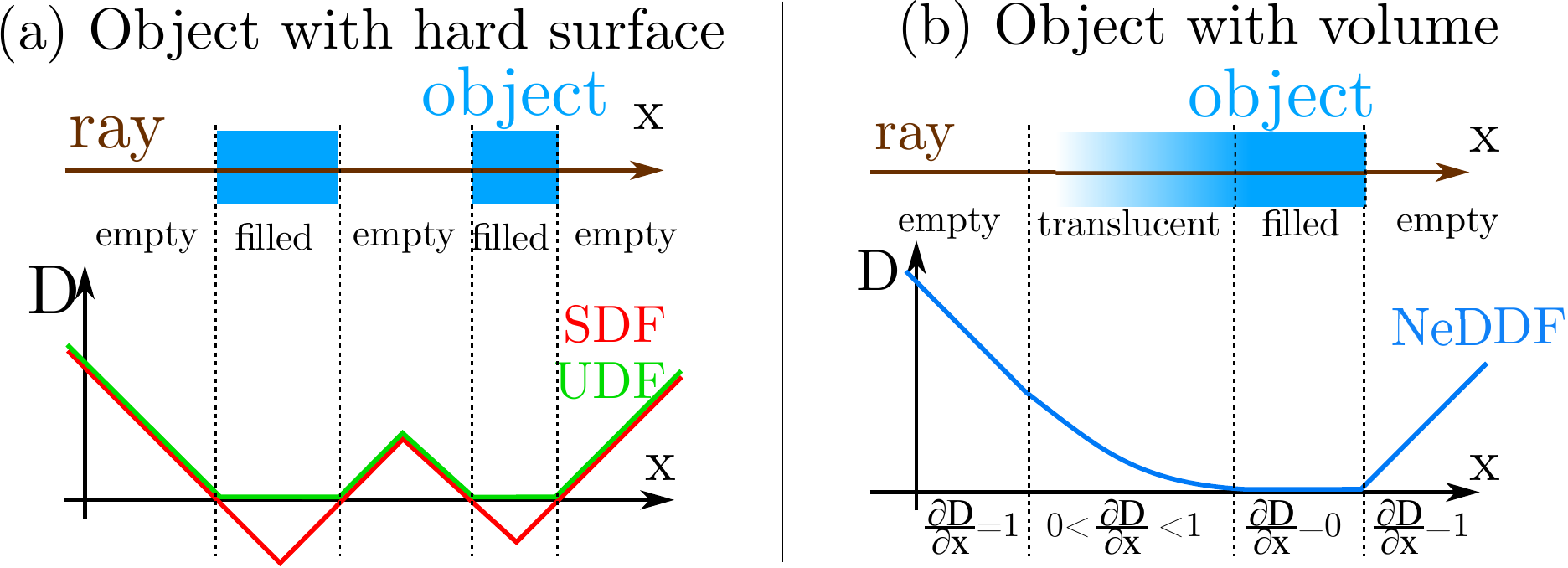}
    \end{center}
    \caption{ (a) Signed Distance Field (SDF) and Unsigned Distance Field (UDF) distinguish the inside and outside of objects by the sign of the distance field ($D>0$ or $D=0$). In addition, UDF does not have a gradient inside the object, but can distinguish between inside and outside of objects by the magnitude of the gradient. (b) NeDDF assigns semitransparent density information to the distance gradient from 0 to 1.}
    \label{fig:ProposedDistField}
\end{figure}

\begin{figure}[ht]
    \begin{center}
    \includegraphics[width=0.6\linewidth]{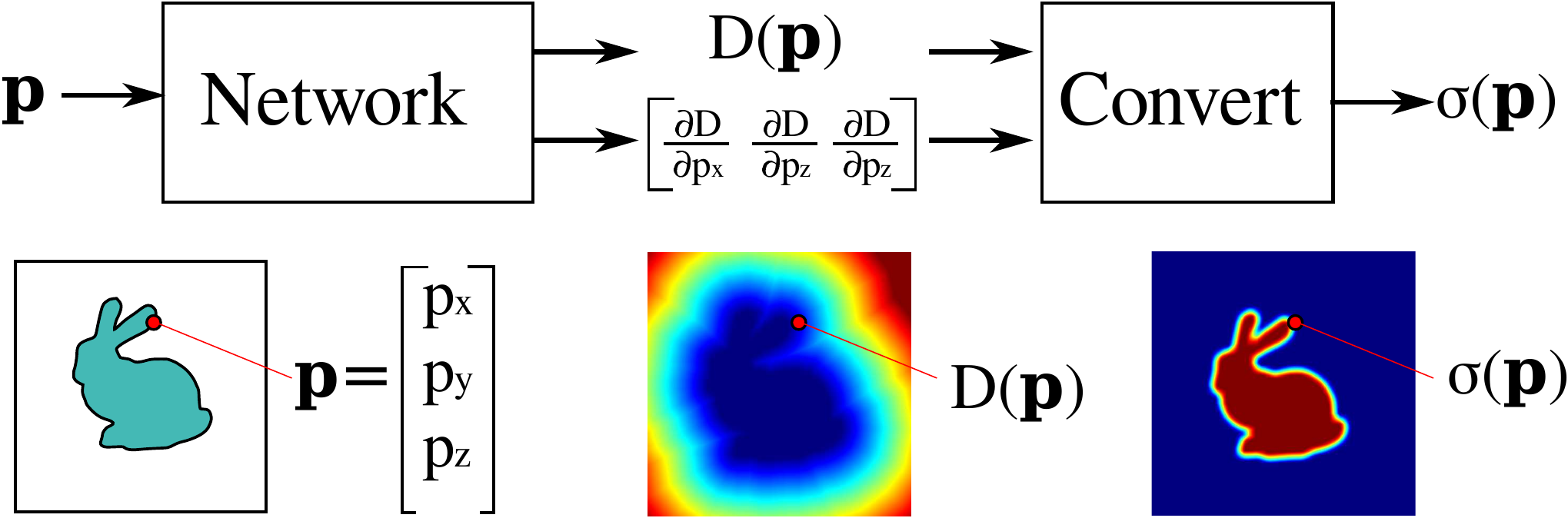}
    \end{center}
    \caption{Flow of proposed model: created network takes the position $\mathbf{p}$ as input and outputs the distance $D$ and its gradient. Differentiable conversion formula for the network output gives the density of position $\sigma(\mathbf{p})$.}
    \label{fig:approach}
\end{figure}

As shown in Fig. \ref{fig:ProposedDistField} (a), we focus on the Unsigned Distance Field (UDF), which ignores surface direction inside objects and can distinguish between the inside and outside objects not only by the sign of distance $D$ but also by the magnitude of its gradient. 
We extend the distance field to be able to recover arbitrary density distributions by interpreting $D$ by the depth derived from the volume rendering equation and fitting the density information of translucent objects to the mid-level gradient magnitude, as shown in Fig. \ref{fig:ProposedDistField} (b).
This method eliminates the need for constraints on the density, as in NeuS, when learning the distance field from images.
In other words, when learning the density field, we can simultaneously obtain a consistent distance field where the shape and camera pose have the same optimal values.
As shown in Fig. \ref{fig:approach}, the NeDDF has a network that inputs a position and outputs the distance and its gradient, and a converter that explicitly calculates the density.
NeDDF enables both high expressiveness of the density field and good registration of the associated distance field.

The present paper provides the following three contributions:
(1) Extending the distance field to be definable for arbitrary density distributions.
(2) Proposing a method to recover the corresponding density from independent points using the distance field and gradient information.
(3) Providing an implementation to alleviate the instability of the distance gradient caused by cusp points and sampling frequency.
Furthermore, the effectiveness of the proposed method in terms of both expressiveness and registration performance is evaluated through experiments.

\section{Related Work}
\subsection{Neural Fields}
The traditional way of representing volumes is to discretize the density or distance from the surface at each position into voxels \cite{kinectfusion,dynamicfusion,splitfusion,li2020learning,lee2020texturefusion,ge20173d}.
Since voxels require data complexity that is cubic in resolution, it is difficult to increase the spatial resolution.
In recent years, memory-efficient representations such as octree~\cite{chen1988survey,SparseVoxelOctree} or hash table~\cite{niessner2013real} have been proposed.
However, these grid structures cannot represent the geometry information with higher frequencies than the Nyquist frequency.
With advances in geometric deep learning following AtlasNet~\cite{AtlasNet} and Foldingnet~\cite{Foldingnet}, some studies have focused on handling irregular non-grid structures such as point clouds and meshes.
While these methods can handle detailed geometry information efficiently, they are limited to interpolation due to their spatially discrete representations based on the coordinates of each point and vertex.

In recent years, the methods called neural fields~\cite{xie2021neuralfield} which directly represent continuous signals by embedding implicit surfaces with neural networks, has been attracting significant interest in the research community.
Given sufficient parameters, fully connected neural networks can encode continuous signals over arbitrary dimensions.
Since a neural field automatically adapts the expressiveness of the network to high frequency regions, it is possible to obtain a shape representation with high resolution using a significantly smaller number of parameters than conventional discrete representation based methods.
Furthermore, since neural fields are continuous representation, we can expand the input and output dimensions without increasing the model capacity.
It is also possible to model topological changes by considering the density field as a three-dimensional slice of high dimensional space and embed time-series information by adding a multi-dimensional deformation code to the input~\cite{HyperNeRF}.

By using a smooth activation function in the neural networks, neural fields can be regarded as a continuously differentiable field.
Modeling using gradient information has been proposed, such as divergence in Non-Rigid-NeRF~\cite{nonrigNeRF} and elasticity constraints using the Jacobian matrix of the deformation field in Nerfies~\cite{Nerfies}.
Inspired by the ideas behind these approaches, we have developed a model in which the gradient of the distance field describes the density information.
The proposed model outputs a distance field and a density field that are explicitly consistent with each other.
Since an optimization-based penalty term does not constrain the model, it can be optimized reciprocally from an objective function that is appropriate for each field.

\subsection{Density Field}
The density field outputs the volume density for the input of the 3D position.
Many methods, such as NeRF, use the density field together with the color field, thus enabling volume rendering.
The density field is characterized by high expressiveness.
For example, a low value of the density field can describe a semi-transparent object such as glass or smoke to represent proportional light transmission.
For spatially high-frequency shapes such as fur, for which the boundary surface is complex, the field can describe a scene by considering the light interaction at an arbitrary point as a function of density, ray direction, and color.
In particular, combining this with the color field makes it possible to model specular reflections, including viewing angle dependency.

Although NeRF can describe complex scenes, it has a substantial limitation in that the camera pose should be known for the observed image and the scene should be static, making it challenging to capture a usable set of images with unknown camera poses.
Therefore, many NeRF-based methods for estimation of camera pose and registration of object deformation have been proposed~\cite{iNeRF,BARF}.
However, since blank areas with a density value of 0 have uncertain gradient directions, camera pose tracking is only valid with initial values such that most of the object area overlaps.
NeRF$--$~\cite{NeRFmm} optimizes camera parameters directly with backpropagation but is limited to camera configurations close to the line of sight.
NSFF~\cite{NSFF} requires the optical flow to follow the deformation.
D-NeRF is limited to CG images with no background and low-frequency texture.
Nerfies~\cite{Nerfies} adds warmup to positional embedding and delays learning of high-frequency components to ensure registration but is limited to camera configurations with close view directions.

NeDDF provides a consistent distance field while retaining the expressiveness of the density field.
By providing gradients where no objects are present, we can improve registration performance from a rough initial camera poses.

\subsection{Distance Field}
A distance field takes a 3D position as input and outputs the distance to the nearest neighbor boundary.
SDF is widely used in fusion and registration because it can provide stable bounding surfaces and normal vectors.
SDF also provides residuals and gradient directions, enabling fast-fitting of two shapes by the Gauss-Newton method without corresponding point matching~\cite{kinectfusion,dynamicfusion,splitfusion,li2020learning,lee2020texturefusion,ge20173d}.
A typical example is KinectFusion~\cite{kinectfusion}, which performs localization and shape integration from a depth map of unknown viewpoints for a static scene.
DynamicFusion~\cite{dynamicfusion} constructs a sparse deformation field called WarpField to describe the deformation amount and performs registration for non-rigid scenes.

In addition, several studies have proposed methods by which to handle distance fields by neural fields.
DeepSDF~\cite{deepsdf} proposes a generative model for the continuous SDFs based on the auto-decoder model.
SAL~\cite{sal} enables the neural field to learn the shapes of boundary surfaces directly from raw unsigned data such as point clouds.
UDF~\cite{UDF} makes the unsigned distance field continuous and shows its suitability for unclosed surfaces and complex shapes.
Since density fields such as NeRF are noisy in surface reconstruction by level sets, several methods have been proposed to handle distance fields in neural fields that can present boundary surfaces and assume a static density distribution for the signed distance.
IDR~\cite{IDR} introduces differentiable surface rendering to learn the neural field from multi-view images.
However, unlike volume rendering, calculating a single surface intersection for each ray makes shape reconstruction unstable for complex shapes that cause abrupt depth changes in the image.
UNISURF~\cite{UNISURF} enables a combination of surface and volume rendering with a neural field that describes occupancy.
Several studies attempt to optimize the SDF with differentiable volume rendering by modeling the transform equation from the distance to the density field with the hypothesis on the shape of the density distribution.
VolSDF~\cite{volSDF} interprets the volume density as Laplace's cumulative distribution function for SDF.
NeuS~\cite{NeuS} assumes a density distribution over the signed distance using the learnable variance values.
However, these distance fields assume the existence of a boundary surface, which limits the kind of shape that can be represented.
This study extends the distance field to correspond to various density distributions from depth values derived from volume rendering.
The present extends the distance field to correspond to various density distributions from depth values derived from volume rendering.

\section{Method}
In this section, we consider reciprocally constrained distance and density fields.
Section 3.1 redefines the distance field to interpret arbitrary density fields, including boundaryless scenes.
Section 3.2 introduces a conversion formula to obtain the density of independent points from the distance and the gradient of the distance value in the redefined distance field.

\subsection{Distance Field from Density Field}
\begin{figure}[ht]
    \begin{center}
    \includegraphics[width=0.2\linewidth]{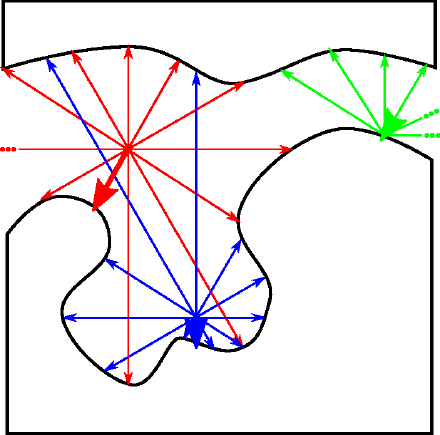}
    \end{center}
    \caption{The distance field uses depth for the nearest surface.}
    \label{fig:MinDist}
\end{figure}

\begin{figure}[ht]
    \begin{center}
    \includegraphics[width=0.7\linewidth]{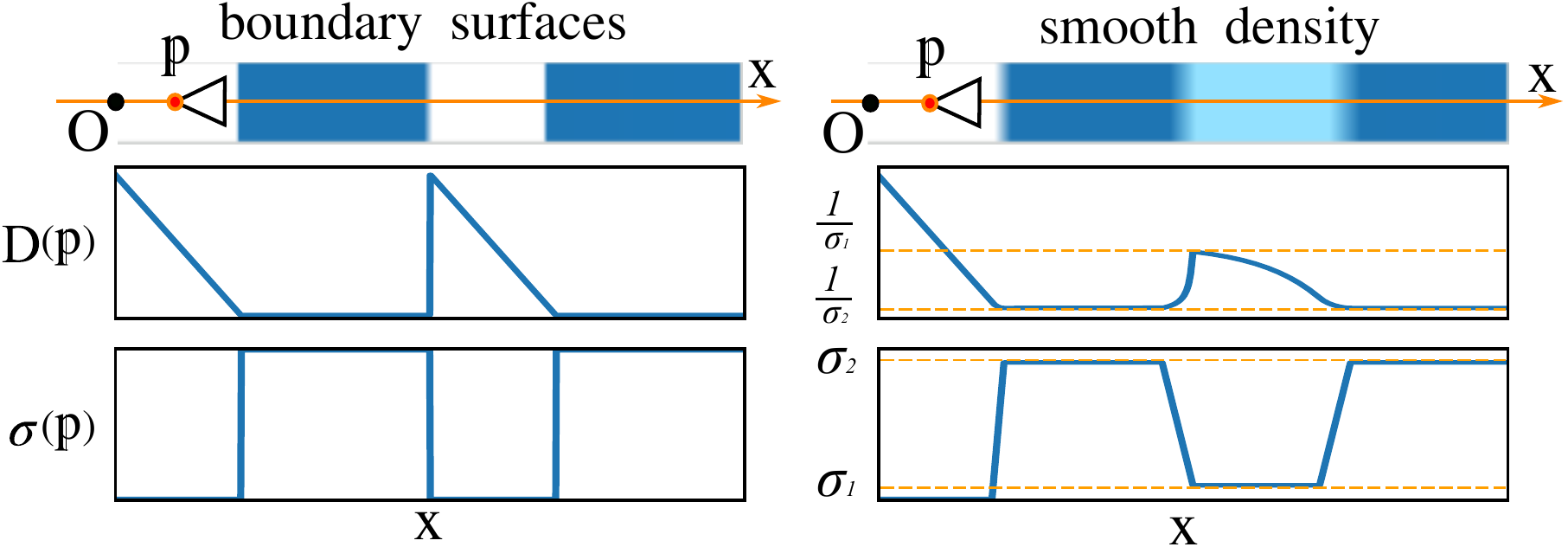}
    \end{center}
    \caption{Depth of volume rendering in case camera is on the x-axis. In a translucent region of density $\sigma$, the upper bound of the depth value becomes $1/\sigma$. When $\sigma$ takes $\infty$ inside the object, the depth value takes $0$, corresponding to the conventional hard surface model.}
    \label{fig:Depth_Bound}
\end{figure}

The distance field in boundary surfaces $D_{b}(\mathbf{p})$ describes the distance to the nearest surface for location $\mathbf{p}\in \mathbb{R}^3$.
We can interpret $D_b(\mathbf{p})$ as the minimum of the depth value $d_b(\mathbf{p},\mathbf{v})$ over the viewing direction $\mathbf{v}\in \mathcal{S}^2$ (see Fig.~\ref{fig:MinDist}):
\begin{equation}
    D_b(\mathbf{p}) := \underset{\mathbf{v} \in S^2}{\mathrm{min}}[d_b(\mathbf{p}, \mathbf{v})].
\end{equation}
We extend the distance field to be defined for arbitrary density distributions by replacing the depth value $d_b(\mathbf{p},\mathbf{v})$ with the depth value derived from the volume rendering equation (see Fig.~\ref{fig:Depth_Bound}).

Assuming that the density distribution $\sigma(\mathbf{p})$ is known, we can calculate the depth value $d(\mathbf{p},\mathbf{v})$ for a viewpoint position $\mathbf{p}$ and a viewing direction $\mathbf{v}$.
Equation \ref{eq:volume_rendering} is the volume rendering formula in NeRF \cite{NeRF}.
For a point on the ray $\mathbf{r}(t)=\mathbf{p}+t \mathbf{v}$ with the visible range $[t_n, t_f]$, the color of each ray $\mathbf{C}(\mathbf{r})$ is obtained through the integral of each color $\mathbf{c}(\mathbf{r}(t),\mathbf{v})$ multiplied by transmission rate $T(t)=\mathrm{exp}\left( -\int_{t_n}^t \sigma(\mathbf{r}(s))ds \right)$:

\begin{equation}
\label{eq:volume_rendering}
  C(\mathbf{r}) := \int_{t_n}^{t_f}T(t)\sigma(\mathbf{r}(t))\mathbf{c}(\mathbf{r}(t), \mathbf{v})dt.
\end{equation}
Similarly, the depth $d(\mathbf{p},\mathbf{v})$ is defined to be  an integral of the depths at each point, as follows:
\begin{equation}
\label{eq:volumetric_depth}
    d(\mathbf{p},\mathbf{v}) := t_n + \int_{t_n}^{t_f}t T(t)\sigma(\mathbf{r}(t)) dt.
\end{equation}
Here, $d(\mathbf{p})$ takes the same value for the depth value $d_b(\mathbf{p})$ in the presence of a boundary surface by taking the density $\sigma(\mathbf{p})$ to be $0$ outside ($0<t<d_b$) and $\infty$ inside ($d_b \leq t$). 

In practice, calculating the depths over all directions is computationally expensive, so we use a distance field that removes the dependence on the viewing direction $\mathbf{v}$.
We define the distance field $D(\mathbf{p}) := \underset{\mathbf{v} \in S^2}{\mathrm{min}}[d(\mathbf{p}, \mathbf{v})]$, for adopting the shortest depth for each position as in the bounding surface model.
Assuming continuity in the adopted view direction $\mathbf{v}_n=\underset{\mathbf{v} \in S^2}{\mathrm{argmin}} [d(\mathbf{p}, \mathbf{v})]$, we can restore this quantity from the tangent plane using the gradient of the distance field $\nabla D(\mathbf{p})$ as follows:
\begin{eqnarray}
\label{eq:view_dir}
    \mathbf{v}_n &=& \frac{-\nabla D(\mathbf{p})}{\|\nabla D(\mathbf{p})\|_2},\\
    \nabla D(\mathbf{p}) &=& 
    \begin{bmatrix}
        \frac{\partial D(\mathbf{p})}{\partial p_x} &
        \frac{\partial D(\mathbf{p})}{\partial p_y} &
        \frac{\partial D(\mathbf{p})}{\partial p_z} 
    \end{bmatrix}.
\end{eqnarray}
In practice, $\mathbf{v}_n$ has discontinuities which makes this calculation difficult. In the next section, we will discuss some strategies to alleviate this.

\subsection{Density from Distance Field}
In Section 3.1, we extended the distance field to shapes with no explicit boundary surface but with a known density field.
In this section, we derive the corresponding density field when the distance field is known.

For the distance field around position $\mathbf{p}$, we consider $D(\mathbf{r}(t)), \mathbf{r}(t)=\mathbf{p}+t\mathbf{v}$, which is sliced in the gradient direction v.
Calculating the derivative of the distance field in the direction of the gradient, $\frac{\partial D(\mathbf{r}(t))}{\partial t}$, we can derive an expression for $\sigma$ as follows: (the derivation is given in supplementary material)
\begin{eqnarray}
    \frac{\partial D(\mathbf{r}(t))}{\partial t}|_{t=0} &=& \lim_{\Delta t \rightarrow 0} \frac{d(\mathbf{r}(\Delta t),\mathbf{v})-d(\mathbf{r}(0),\mathbf{v})}{\Delta t}\\
    \label{eq:dist_density}
    &=& -1 + (D(\mathbf{p})-t_n)\sigma(\mathbf{p}+t_n \mathbf{v}).
\end{eqnarray}
We can also express $\frac{\partial D(\mathbf{r}(t))}{\partial t}$ using the gradient vector of the distance field $\nabla D(\mathbf{p})$ as follows:
\begin{eqnarray}
    \frac{\partial D(\mathbf{r}(t))}{\partial t} &=& 
    \frac{\partial D}{\partial p_x}  \frac{\partial p_x}{\partial t} + 
    \frac{\partial D}{\partial p_y}  \frac{\partial p_y}{\partial t} + 
    \frac{\partial D}{\partial p_z}  \frac{\partial p_z}{\partial t}\\
    &=& \nabla D(\mathbf{p}) \cdot \mathbf{v}\\
    \label{eq:dist_grad}
    &=& -\|\nabla D(\mathbf{p})\|_2.
\end{eqnarray}
From Equations \ref{eq:dist_density} and \ref{eq:dist_grad}, the density $\sigma$ can be obtained as follows:
\begin{equation}
    \sigma(\mathbf{p}) = \frac{1-\|\nabla D(\mathbf{p}-t_n \mathbf{v})\|_2}{D(\mathbf{p}-t_n \mathbf{v})-t_n}.
\end{equation}
From Equation \ref{eq:volumetric_depth}, $t_n$ is the interval in which the transmittance $T(t_n)=1$, which is the lower limit of $D$.
While it is natural for $t_n$ to take the value 0, the density is undefined for $D\sim 0$ as a numerical problem.
Assuming $t_n$ is sufficiently small, by introducing the approximations in Equations \ref{eq:sim_dpt},\ref{eq:sim_gpt},  $\sigma$ can be calculated as in Equation \ref{eq:convert}:
\begin{eqnarray}
\label{eq:sim_dpt}
D(\mathbf{p}-t_n \mathbf{v})&\simeq& D(\mathbf{p})+t_n,\\
\label{eq:sim_gpt}
\nabla D(\mathbf{p}-t_n \mathbf{v}) &\simeq& \nabla D(\mathbf{p}),\\
\label{eq:convert}
    \sigma(\mathbf{p}) &\simeq& \frac{1-\|\nabla D(\mathbf{p})\|_2}{D(\mathbf{p})}.
\end{eqnarray}
The differentiability of the network allows us to determine the distance $D(\mathbf{p}_i)$ and gradient vector $\nabla D(\mathbf{p}_i)$ with independent sampling points $\mathbf{p}_i$ as input to the neural fields, such as regressing a distance field.
Equation \ref{eq:convert} allows us to compute the density $\sigma(\mathbf{p}_i)$ with a differentiable conversion formula.
In other words, it is possible to learn mutually constrained distance and density fields.
Note that from $t_n>0$, the density is limited to $\sigma(\mathbf{p})<\frac{1}{t_n}$.

\subsection{Removing Cusps}
\begin{figure}[ht]
    \begin{center}
    \includegraphics[width=0.9\linewidth]{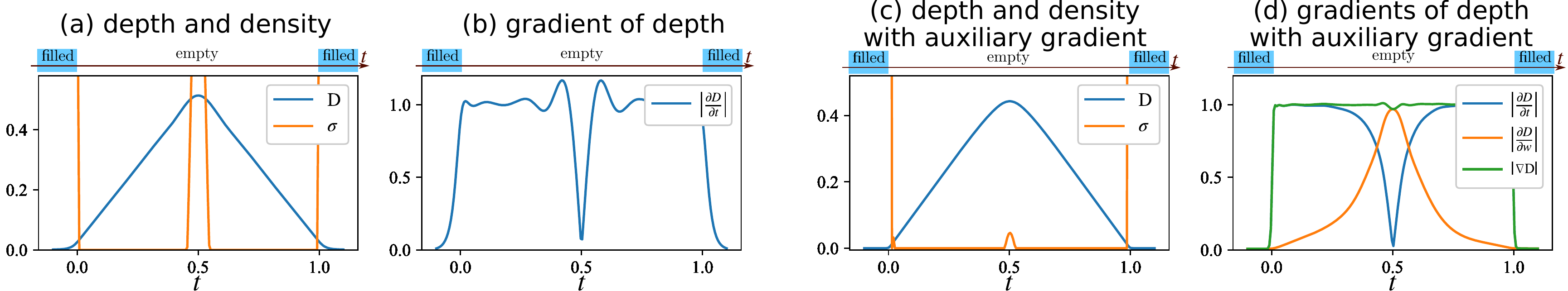}
    \end{center}
    \caption{
    A visualization of each field in the toy model with $0\leq t \leq 1$ as object exterior. 
    }
    \label{fig:wdim_wo}
\end{figure}

The distance to density conversion by Equation \ref{eq:convert} assumed that the distance field is first-order differentiable. However, a distance field is not differentiable at the cusps where the minimum distance direction switches. In practice, although the distance field around the cups is smoothly connected due to the continuity of the neural field, small gradient values around the cups are still converted to false densities that should not exist by Equation \ref{eq:convert} (see Fig.~\ref{fig:wdim_wo}). To alleviate this problem at the caps, we extend the domain of the distance field from a 3-dimensional space to a 4-dimensional hyperspace $[x,y,z,w]$ with an auxiliary gradient axis ($w$-axis) and consider the slice at $w=0$ of the hyperspace to be the distance field. The gradient components $\frac{\partial D}{\partial w}$ are distributed so that the gradient $\nabla D$ satisfies the Equation \ref{eq:convert} in the vicinity of the cusp to suppress spurious densities.

When the auxiliary gradient also describes the density outside the cusp region, the shape of the global distance field is compromised.
Therefore, we constrain the shape of the auxiliary gradient and use a penalty term to induce convergence to the optimal form.
In this study, when $\frac{\partial D}{\partial t}>0$, we use the following heuristic constraint:
\begin{equation}
\label{eq:dDdwShape}
    \frac{\partial^2 D}{\partial t \partial w} = \alpha \frac{1}{D} \frac{\partial D}{\partial w}.
\end{equation}
Note that $\alpha$ is a hyperparameter that determines the scale of the gradient.
The shape of the auxiliary gradient for each $\alpha$ is shown in appendix. Since the Equation \ref{eq:dDdwShape} is a constraint other than $\frac{\partial D}{\partial t}=0$, and since it becomes unstable around $D=0$ and $\frac{\partial D}{\partial w}=0$, we introduce the following weight coefficient $\beta$:
\begin{equation}
    \beta = D \left(\frac{\partial D}{\partial t}\right)^2 \frac{\partial D}{\partial w}.
\end{equation}
Note that since $\beta$ is a value for discriminating the target point, it is separated from the calculation graph during training like stop-gradient operator.
For M sampling points $\mathcal{P}$, the objective function $L_{\mathrm{const}}$ for the shape constraint of the auxiliary gradient is set as:
\begin{equation}
    L_{\mathrm{const}} = \frac{\lambda_{\mathrm{const}}}{M} \sum_{\mathbf{p} \in \mathcal{P}} \beta \left[\frac{\partial^2 D}{\partial t \partial w} - \frac{\alpha}{D(\mathbf{p})} \frac{\partial D}{\partial w}\right]^2.
\end{equation}
Note that $\lambda_{\mathrm{const}}$ is a hyperparameter.

\section{Reprojection error for volume rendering}
Previous NeRF-based localization such as iNeRF~\cite{iNeRF} uses photometric error, the residual $\|\mathbf{C}(\mathbf{q})-\hat{\mathbf{C}}(\mathbf{q})\|_2$ from the observed color $\mathbf{C}(\mathbf{q})$, and the estimated color $\hat{\mathbf{C}}(\mathbf{q})$ aggregated by volume rendering for selected pixel $\mathbf{q}$.
While photometric error can follow objects without hard surfaces, it can only follow local regions with smooth color changes.
NeDDF provides the object's direction and approximate distance from a sampling point.
Therefore, we can calculate the pseudo-correspondence point and estimate the camera pose using the reprojection error.
This section describes a method that uses color information to calculate the correspondence points as a simple example.

\begin{figure}[ht]
    \begin{center}
    \includegraphics[width=0.8\linewidth]{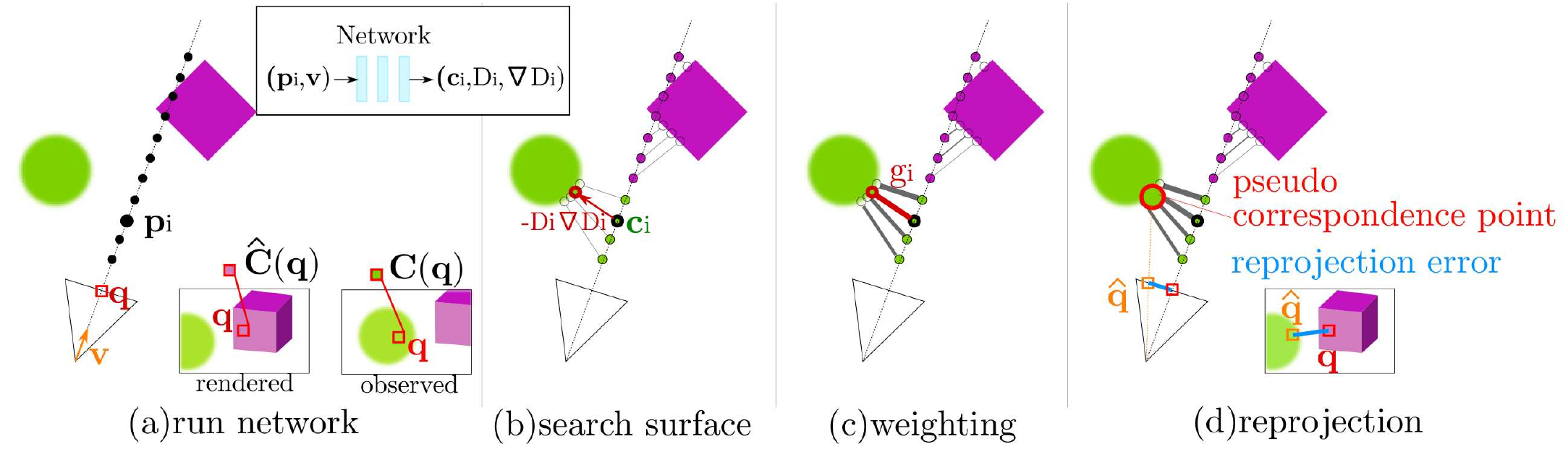}
    \end{center}
    \caption{
    Localization by reprojection error: (a) Network execution outputs the color and distance of each sampling point. Network provides the color $\mathbf{c}_i$, distance $D_i$ and its gradient $\nabla D_i$ for the sampling point $\mathbf{p}_i$.(b) Network output gives the points near the surface. (c) Weight $g_i$ takes the value that emphasizes sampling points with similar colors or smaller distances. (d) Pseudo-correspondence points synthesized from the weights produce the reprojection error.
    }
    \label{fig:reprojection}
\end{figure}

When training the network, the objective function with volume rendering has no constraints about the color field in the blank region. We record the same color in the gradient direction of the distance field by penalizing the color change in the distance gradient direction for a blank region.
For a point $\mathbf{p}_i$, camera depth $t_i$ and a viewing direction $\mathbf{v}$, let the output be color $\mathbf{c}_i$, distance $D_i$, and distance gradient $\nabla D_i$. The penalty takes $L_{\mathrm{blank}} = \sum_i \|\nabla \mathbf{c}_i (\nabla D_i)^T\|_2$. Since $\|\nabla D_i\|_2$ takes small values inside the object, this penalty becomes active for regions outside the object. Training the network by introducing this penalty makes it possible to obtain the nearest neighbor object's color, direction, and distance from sampling points outside the object.

In estimating camera poses, finding the pseudo-correspondence points for each ray provides the calculation of the reprojection error, as shown in the Fig. \ref{fig:reprojection}.
The network output for sampling point $\mathbf{p}_i$ leads to the near-surface point of color $\mathbf{c}_i$ at position $\mathbf{p}_i - D_i(\nabla D_i)$.
We calculate the pseudo-correspondence points for each ray by selecting points closer to $\mathbf{C}(\mathbf{q})$ than $\hat{\mathbf{C}}(\mathbf{q})$ with combining near-surface points by focusing weights closer to the color and the distance between them.
The weight $g_i$ of sampling point $\mathbf{p}_i$ is calculated as follows:
\begin{equation}
g_i = \mathrm{softmax}(-\lambda_D \frac{D_i  \|\nabla D_i \times \mathbf{v}\|_2}{t_i} - \lambda_c \|\mathbf{C}(\mathbf{q}) - \mathbf{c}_i\|_2).
\end{equation}
Note that $\lambda_D$ and $\lambda_c$ are hyperparameters, softmax aggregates for axis with index i.
The reprojection error measures the distance $\|\mathbf{q}-\hat{\mathbf{q}}\|_2$ between the pixel coordinates of the ray $\mathbf{q}$ and the projected pseudo-correspondence point $\hat{\mathbf{q}}$.

\section{Experiments}
\subsection{Experimental setup}
\subsubsection{(a) Quality of novel view synthesis.} 
We confirm that NeDDF retains the comparable quality of novel view synthesis as NeRF.
We use the NeRF synthetic dataset which contains CG images of the eight scenes rendered from camera positions placed on a hemispherical surface.
Using NeRF and NeuS as baselines, we compare the quality of the new viewpoint images using PSNR.

\subsubsection{(b) Accuracy of localization.} 
In NeDDF, the pseudo correspondence points enable estimation of camera pose using reprojection error.
We confirm that the use of reprojection error improves the accuracy of localization from only using photometric error under poor initial camera poses.
The experiment uses 200 test viewpoints of the Lego scene in the NeRF synthetic Dataset.
Each camera takes the initial pose given by random rotations and translations applied to the ground truth values of the test viewpoint.
We evaluate the camera position and angle errors for three cases: optimizing the camera pose for 300 iterations by photometric error, 300 iterations by reprojection error, and 100 iterations by reprojection error plus 200 iterations by photometric error.
The localization flows follow similar practices to iNeRF.
The optimization uses the gradient of the camera pose as a 6D parameter of SE(3) and sets the increment by Adam~\cite{kingma2015adam} with exponential decay.
In each iterations, interest region sampling~\cite{iNeRF} selects 256 rays for optimization.

\subsection{Results}
\subsubsection{(a) Quality of novel view synthesis.} 
\begin{figure}[ht]
    \begin{center}
    \includegraphics[width=0.9\linewidth]{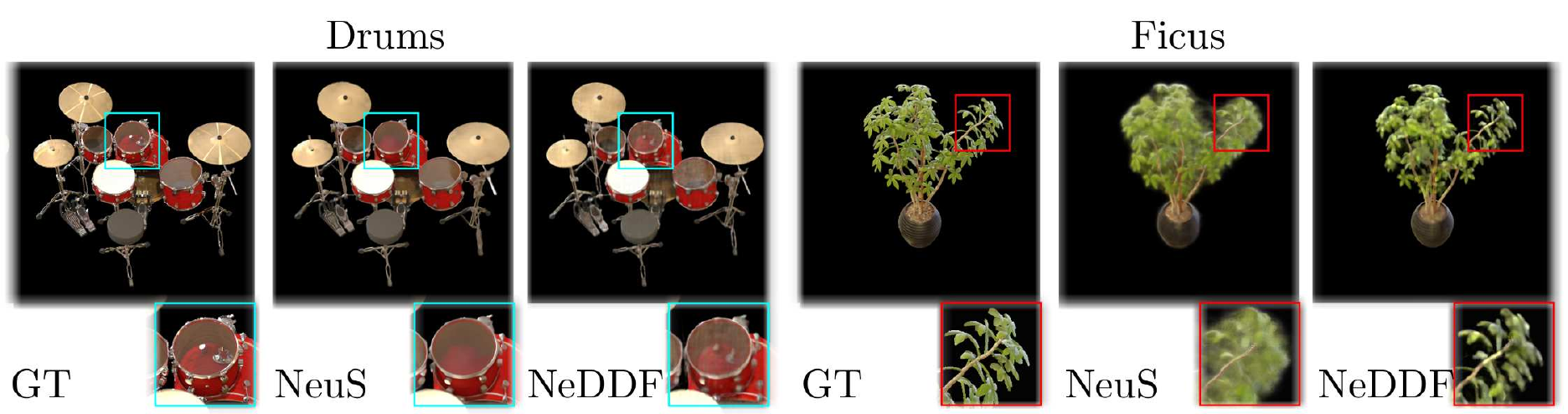}
    \end{center}
    \caption{Comparisons on test-set views for scenes from NeRF synthetic dataset generated with a volume renderer.}
    \label{fig:result_render}
\end{figure}

Table~\ref{table:result_psnr_synthetic} shows the PSNR for each scene and each method for evaluating the quality of the generated images
The NeDDF results retain the comparable quality of the state-of-the-art methods in novel view synthesis, although the PSNR comparison is slightly inferior.
Fig.~\ref{fig:result_render} shows visualizations of Drums and Ficus as examples of scenes with transparency and delicate shapes, where conventional distance-field-based methods are weak.
For the Drum scene, NeuS could not represent the transparent parts of the drums and colored the transparent surfaces. In contrast, NeDDF reproduces the transparent parts and obtains the metal parts on the back through.
The NeuS results show blurred images for the Ficus scene since it fails to assign the appropriate density distributions due to the difficulty of understanding the model for thin surfaces such as leaves.
NeDDF provides comparatively high-quality image restoration even for delicate shapes such as leaves.
NeDDF allows for the reconstruction of thinner surfaces than the sampling interval by interpreting their occupancy as a lower density.

\begin{table}[t]
  \centering
  \caption{Quantitative evaluation on synthetic dataset. We report PSNR (higher is better).}
  \begin{tabular}{@{}l|cccccccc|c@{}}
    \toprule
    Method & Chair & Drums & Ficus & Hotdog & Lego & Materials & Mic & Ship & Mean\\
    \midrule
    NeRF & 33.00 & 25.01 & 30.13 & 36.18 & 32.54 & 29.62 & 32.91 & 28.65 & 31.01\\
    NeuS & 27.69 & 22.14 & 21.67 & 32.14 & 27.18 & 25.64 & 27.52 & 23.47 & 25.93\\
    NeDDF & 29.11 & 23.96 & 25.72 & 30.85 & 27.93 & 25.52 & 29.34 & 23.69 & 27.02\\
    \bottomrule
  \end{tabular}
  \label{table:result_psnr_synthetic}
\end{table}

\begin{figure}[ht]
    \begin{center}
    \includegraphics[width=0.5\linewidth]{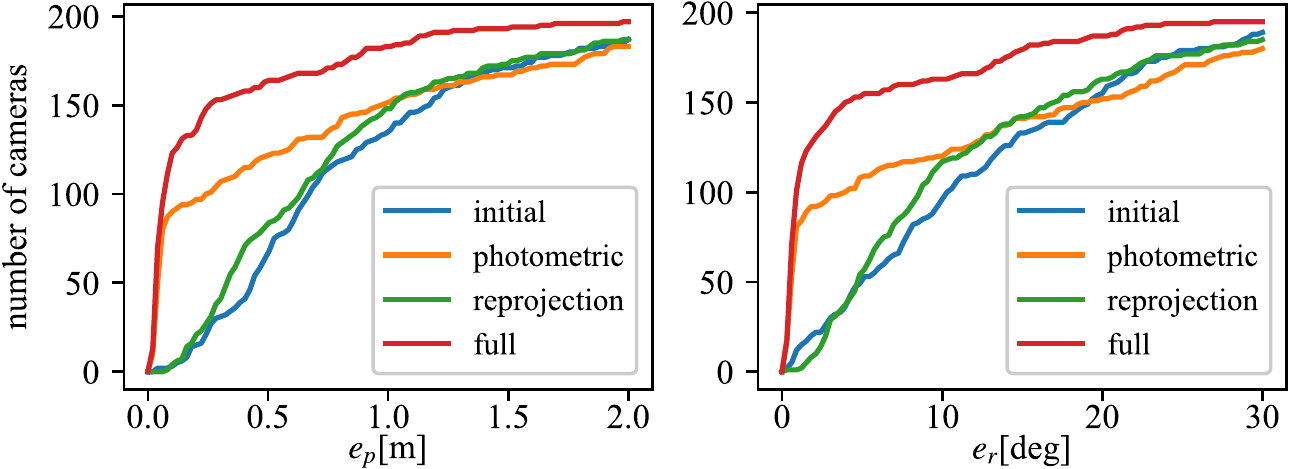}
    \end{center}
    \caption{Quantitative evaluation of camera poses estimation accuracy. The horizontal axis represents the position and angle error, and the vertical axis represents the number of cameras recovered under the errors.}
    \label{fig:result_camera_count}
\end{figure}

\begin{figure}[ht]
    \begin{center}
    \includegraphics[width=0.9\linewidth]{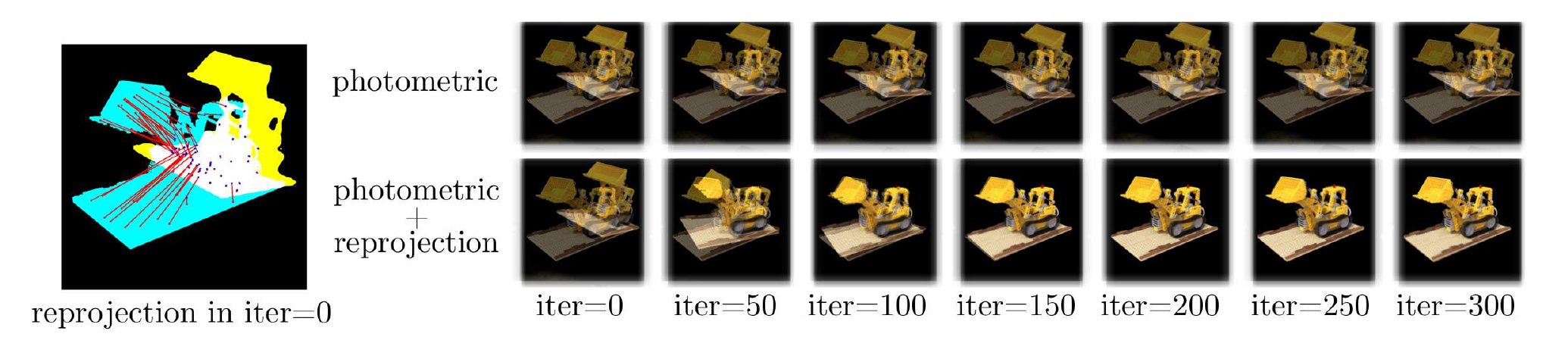}
    \end{center}
    \caption{ (left) Projection of the corresponding points in iteration 0: the cyan region is the rendered object region of the observed image, and the yellow region is the rendered object region estimated from the current camera attitude. The red dots denote sampling points, and the blue dots denote the projected positions of the calculated correspondence points. (right) Observed and rendered images at each optimization iteration: the upper row shows the case optimized using only photometric error, and the lower row shows the case optimized using the photometric and reprojection errors.}
    \label{fig:result_tracking}
\end{figure}

\subsubsection{(b) Accuracy of localization.}
Fig.~\ref{fig:result_camera_count} shows a plot of the camera attitude estimation results.
Optimization by photometric error works very accurately when the initial value of the position error is approximately 0.5 [m] or less, and the initial value of the angular error is 10 [deg] or less, while the error increases when the initial value of the error is significant.
Since photometric error only provides reasonable gradient directions in smooth color changing, significant initial value errors may lead to erroneous local solutions.
Optimization by reprojection error does not improve camera posture residuals much on its own.
This is because the correspondence point calculation based on color similarity lacks uniqueness, which leads to many mismatches in the correspondence points.
On the other hand, optimization by reprojection error can roughly estimate the camera pose such that regions with close colors overlap. Using this as a preprocessing step for optimization by photometric error can significantly expand the range of initial values for highly accurate estimation.

Fig.~\ref{fig:result_tracking} shows the localization process with large initial value errors.
 The left figure shows the projected positions of the sampling points and the calculated correspondence points at iter=0.
The figure shows that even when the sampling points are far from the object, as in the lower-left region, we can obtain the corresponding points in the object region.
On the other hand, in the upper left area of the bucket, the model selects the wrong correspondence points.
 We can reduce the impact of such rays by reducing the weights or calculating the corresponding points using features with higher uniqueness instead of color.
The figure on the right shows the overlaid rendered and observed images at each optimization iteration.
The overlapped object area between the rendered and observed images is small for iter=0.
Optimization using only photometric error gives little progress since it cannot provide a reasonable gradient.
Rough alignment using the reprojection error by the 100th iteration enables the use of photometric error with an effective gradient, which provides highly accurate localization.

\section{Conclusions}
This study proposed NeDDF to represent reciprocally constrained distance and density fields.
We extended the distance field to a formulation that can adapt to any density field.
We also derived the conversion formula between distance and density using the distance and its gradient, enabling learning the these fields while constraining each other.
We also alleviate the problem of discontinuity points by introducing the auxiliary gradient.
The visualization experiments demonstrated that NeDDF could acquire the properties of both the conventional density field and the distance field.
The quantitative evaluation showed that NeDDF could provide competitive quality of novel view synthesis, more stable meshes, and a more comprehensive range of following camera poses than NeRF.

One limitation of our method is the lack of information about the distance field inside the objects. 
Since NeDDF is based on UDF, it cannot provide helpful gradient directions in the interior object region.
NeDDF also has the same limitations as the original NeRF~\cite{NeRF}, such as time-consuming optimization and rendering.
However, since NeDDF provides a differentiable density field and retains the same formulation as NeRF, many of the latest advances in improving NeRF, such as speedup and stabilization, may apply to NeDDF.
In addition, we calculated the pseudo correspondence points from the colors in the localization as a simple example of obtaining the reprojection error.
We believe that using information with higher uniqueness, such as semantic segmentation, will increase the usefulness of the reprojection error.

\section*{Acknowledgments}
This work was supported by JSPS KAKENHI Grant Number JP19H00806, JP21KK0070 and JP22H01580.
We would like to thank Mr. Towaki Takikawa, Mr. Naoya Chiba, and Mr. Ryota Suzuki for their helpful discussions.


\bibliographystyle{splncs04}
\bibliography{egbib}

\begin{thebibliography}{10}
\providecommand{\url}[1]{\texttt{#1}}
\providecommand{\urlprefix}{URL }
\providecommand{\doi}[1]{https://doi.org/#1}

\bibitem{sal}
Atzmon, M., Lipman, Y.: Sal: Sign agnostic learning of shapes from raw data.
  In: Proceedings of the IEEE Conference on Computer Vision and Pattern
  Recognition (CVPR). pp. 2565--2574 (2020)

\bibitem{chen1988survey}
Chen, H.H., Huang, T.S.: A survey of construction and manipulation of octrees.
  Computer Vision, Graphics, and Image Processing  \textbf{43}(3),  409--431
  (1988)

\bibitem{UDF}
Chibane, J., Mir, A., Pons-Moll, G.: Neural unsigned distance fields for
  implicit function learning. In: Advances in Neural Information Processing
  Systems (NeurIPS). pp. 21638--21652 (2020)

\bibitem{kingma2015adam}
Diederik P.~Kingma, J.B.: Adam: A method for stochastic optimization. In:
  International Conference on Learning Representations (2015)

\bibitem{ge20173d}
Ge, L., Liang, H., Yuan, J., Thalmann, D.: 3d convolutional neural networks for
  efficient and robust hand pose estimation from single depth images. In:
  Proceedings of the IEEE Conference on Computer Vision and Pattern Recognition
  (CVPR). pp. 1991--2000 (2017)

\bibitem{AtlasNet}
Groueix, T., Fisher, M., Kim, V.G., Russell, B., Aubry, M.: {AtlasNet: A
  Papier-M\^ach\'e Approach to Learning 3D Surface Generation}. In: Proceedings
  of the IEEE Conference on Computer Vision and Pattern Recognition (CVPR). pp.
  216--224 (2018)

\bibitem{SparseVoxelOctree}
Laine, S., Karras, T.: Efficient sparse voxel octrees. IEEE Transactions on
  Visualization and Computer Graphics  \textbf{17}(8),  1048--1059 (2010)

\bibitem{lee2020texturefusion}
Lee, J.H., Ha, H., Dong, Y., Tong, X., Kim, M.H.: Texturefusion: High-quality
  texture acquisition for real-time rgb-d scanning. In: Proceedings of the IEEE
  Conference on Computer Vision and Pattern Recognition (CVPR). pp. 1272--1280
  (2020)

\bibitem{li2020learning}
Li, Y., Bozic, A., Zhang, T., Ji, Y., Harada, T., Nie{\ss}ner, M.: Learning to
  optimize non-rigid tracking. In: Proceedings of the IEEE Conference on
  Computer Vision and Pattern Recognition (CVPR). pp. 4910--4918 (2020)

\bibitem{splitfusion}
Li, Y., Zhang, T., Nakamura, Y., Harada, T.: Splitfusion: Simultaneous tracking
  and mapping for non-rigid scenes. In: IROS. pp. 5128--5134 (2020)

\bibitem{NSFF}
Li, Z., Niklaus, S., Snavely, N., Wang, O.: Neural scene flow fields for
  space-time view synthesis of dynamic scenes. In: Proceedings of the IEEE
  Conference on Computer Vision and Pattern Recognition (CVPR). pp. 6498--6508
  (2021)

\bibitem{BARF}
Lin, C.H., Ma, W.C., Torralba, A., Lucey, S.: Barf: Bundle-adjusting neural
  radiance fields. In: Proceedings of the IEEE International Conference on
  Computer Vision (ICCV). pp. 5741--5751 (2021)

\bibitem{tanhexp}
Liu, X., Di, X.: Tanhexp: A smooth activation function with high convergence
  speed for lightweight neural networks. arXiv preprint arXiv:2003.09855
  (2020)

\bibitem{NeRF}
Mildenhall, B., Srinivasan, P.P., Tancik, M., Barron, J.T., Ramamoorthi, R.,
  Ng, R.: Nerf: Representing scenes as neural radiance fields for view
  synthesis. In: Proceedings of the European Conference on Computer Vision
  (ECCV). pp. 405--421 (2020)

\bibitem{dynamicfusion}
Newcombe, R.A., Fox, D., Seitz, S.M.: Dynamicfusion: Reconstruction and
  tracking of non-rigid scenes in real-time. In: Proceedings of the IEEE
  Conference on Computer Vision and Pattern Recognition (CVPR). pp. 343--352
  (2015)

\bibitem{kinectfusion}
Newcombe, R.A., Izadi, S., Hilliges, O., Molyneaux, D., Kim, D., Davison, A.J.,
  Kohi, P., Shotton, J., Hodges, S., Fitzgibbon, A.: Kinectfusion: Real-time
  dense surface mapping and tracking. In: 10th IEEE International Symposium on
  Mixed and Augmented Reality (ISMAR). pp. 127--136 (2011)

\bibitem{niessner2013real}
Nie{\ss}ner, M., Zollh{\"o}fer, M., Izadi, S., Stamminger, M.: Real-time 3d
  reconstruction at scale using voxel hashing. ACM Transactions on Graphics
  (ToG)  \textbf{32}(6),  1--11 (2013)

\bibitem{UNISURF}
Oechsle, M., Peng, S., Geiger, A.: Unisurf: Unifying neural implicit surfaces
  and radiance fields for multi-view reconstruction. In: Proceedings of the
  IEEE International Conference on Computer Vision (ICCV). pp. 5589--5599
  (2021)

\bibitem{deepsdf}
Park, J.J., Florence, P., Straub, J., Newcombe, R., Lovegrove, S.: Deepsdf:
  Learning continuous signed distance functions for shape representation. In:
  Proceedings of the IEEE Conference on Computer Vision and Pattern Recognition
  (CVPR). pp. 165--174 (2019)

\bibitem{Nerfies}
Park, K., Sinha, U., Barron, J.T., Bouaziz, S., Goldman, D.B., Seitz, S.M.,
  Martin-Brualla, R.: Nerfies: Deformable neural radiance fields. Proceedings
  of the IEEE International Conference on Computer Vision (ICCV) pp. 5865--5874
  (2021)

\bibitem{HyperNeRF}
Park, K., Sinha, U., Hedman, P., Barron, J.T., Bouaziz, S., Goldman, D.B.,
  Martin-Brualla, R., Seitz, S.M.: Hypernerf: A higher-dimensional
  representation for topologically varying neural radiance fields. ACM
  Transactions on Graphics (ToG)  \textbf{40}(6),  1--12 (2021)

\bibitem{DNeRF}
Pumarola, A., Corona, E., Pons-Moll, G., Moreno-Noguer, F.: D-nerf: Neural
  radiance fields for dynamic scenes. In: Proceedings of the IEEE Conference on
  Computer Vision and Pattern Recognition (CVPR). pp. 10318--10327 (2021)

\bibitem{nonrigNeRF}
Tretschk, E., Tewari, A., Golyanik, V., Zollh\"{o}fer, M., Lassner, C.,
  Theobalt, C.: Non-rigid neural radiance fields: Reconstruction and novel view
  synthesis of a dynamic scene from monocular video. In: Proceedings of the
  IEEE International Conference on Computer Vision (ICCV). pp. 12959--12970
  (2021)

\bibitem{NeuS}
Wang, P., Liu, L., Liu, Y., Theobalt, C., Komura, T., Wang, W.: Neu{S}:
  Learning neural implicit surfaces by volume rendering for multi-view
  reconstruction. In: Advances in Neural Information Processing Systems
  (NeurIPS) (2021)

\bibitem{NeRFmm}
Wang, Z., Wu, S., Xie, W., Chen, M., Prisacariu, V.A.: Ne{RF}$--$: Neural
  radiance fields without known camera parameters. arXiv preprint arXiv:
  2102.07064  (2021)

\bibitem{xie2021neuralfield}
Xie, Y., Takikawa, T., Saito, S., Litany, O., Yan, S., Khan, N., Tombari, F.,
  Tompkin, J., Sitzmann, V., Sridhar, S.: Neural fields in visual computing and
  beyond. arXiv preprint arXiv: 2111.11426  (2021)

\bibitem{Foldingnet}
Yang, Y., Feng, C., Shen, Y., Tian, D.: Foldingnet: Point cloud auto-encoder
  via deep grid deformation. In: Proceedings of the IEEE Conference on Computer
  Vision and Pattern Recognition (CVPR). pp. 206--215 (2018)

\bibitem{volSDF}
Yariv, L., Gu, J., Kasten, Y., Lipman, Y.: Volume rendering of neural implicit
  surfaces. In: Advances in Neural Information Processing Systems (NeurIPS)
  (2021)

\bibitem{IDR}
Yariv, L., Kasten, Y., Moran, D., Galun, M., Atzmon, M., Ronen, B., Lipman, Y.:
  Multiview neural surface reconstruction by disentangling geometry and
  appearance. In: Advances in Neural Information Processing Systems (NeurIPS).
  vol.~33 (2020)

\bibitem{iNeRF}
Yen-Chen, L., Florence, P., Barron, J.T., Rodriguez, A., Isola, P., Lin, T.Y.:
  {iNeRF}: Inverting neural radiance fields for pose estimation. In:
  Proceedings of the IEEE/RSJ Conference on Intelligent Robots and Systems
  (IROS). pp. 1323--1330 (2021)

\end{thebibliography}

\newpage
\begin{center}
\vspace{4.5cm}
{\LARGE \bf{Appendix}} \\
\vspace{0.5cm}
\end{center}

\section{Architecture details}
\begin{figure}[ht]
    \begin{center}
     \includegraphics[width=1.0\linewidth]{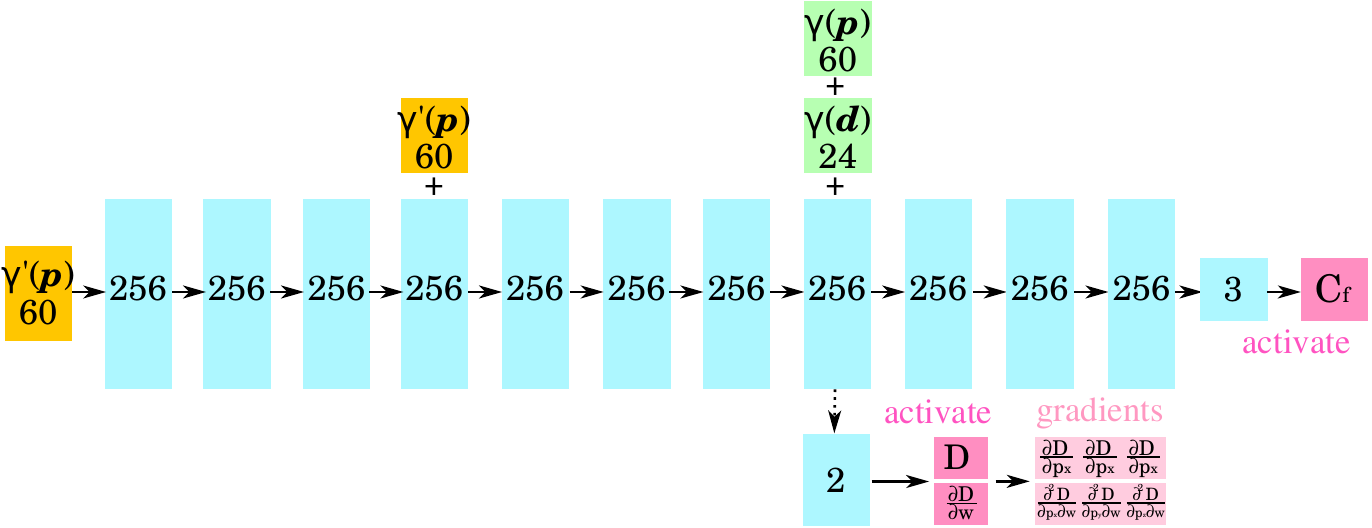}
     \end{center}
     \caption{Details of network architecture in NeDDF}
     \label{fig:network}
\end{figure}

Fig. \ref{fig:network} shows the details of our architecture.
The network uses multilayer perceptrons (MLPs) with a width of 256, the same as in NeRF~\cite{NeRF} and NeuS~\cite{NeuS}.
The network calculates the density using the first derivative value for the output distance $D$.
This calculation requires careful setup of the Positional Encoding (PE) and activation functions.
Using an objective function for the density field requires gradients up to the second derivative for activation functions.
The architecture uses tanhExp~\cite{tanhexp} as the activation function whose second derivative is continuous.
In the conventional method~\cite{NeRF}, the PE up to $L$ dimensions utilizes values such as the following equation:
\begin{equation}
    \gamma({\mathbf{p}}) = \left[ 
    \mathrm{sin}(\mathbf{p}),\mathrm{cos}(\mathbf{p}),
    \cdots,
    \mathrm{sin}(2^{L-1}\mathbf{p}),\mathrm{cos}(2^{L-1}\mathbf{p})
    \right]^T.
\end{equation}

It is the concatenation of the $\mathrm{sin}$ and $\mathrm{cos}$ of each dimension of position $\mathbf{p}$ scaled by powers of 2 from $1$ to $2L-1$.
This PE amplifies by the frequency in the hierarchy of first derivatives, which emphasizes the high-frequency elements in the density field.
In other words, the maximum and minimum frequency components have a scale difference of $2^{L-1}$ in their influence on the density derivative, thus making the learning process unstable.
Therefore, our architecture damps the high-frequency element so that the scale in the single-differentiation hierarchy is the same as the original PE, as shown in the following equation:
\begin{equation}
    \gamma'({\mathbf{p}}) = \left[ 
    \mathrm{sin}(\mathbf{p}),\mathrm{cos}(\mathbf{p}),
    \cdots,
    \frac{1}{2^{L-1}}\mathrm{sin}(2^{L-1}\mathbf{p}), \frac{1}{2^{L-1}}\mathrm{cos}(2^{L-1}\mathbf{p})
    \right]^T.
\end{equation}

However, since the PE neglects the high-frequency component in the non-derivative hierarchy, we need to add an intermediate input of the conventional $\gamma(\mathbf{p})$ in the layers after the distance output for learning detailed color fields.
Note that the performance of restoring the high-frequency component of the color field is worse than that of NeRF for the same network size.

\section{Deriving the conversion from distance to density}
This section describes the derivation details of Equation 10 in the main paper.
For the distance field around position $\mathbf{p}\in \mathbb{R}^3$, we consider $D(\mathbf{r}(t)), \mathbf{r}(t)=\mathbf{p}+t\mathbf{v}$, which is sliced in the gradient direction $\mathbf{v}$.
Calculating the derivative of the distance field in the direction of the gradient, $\frac{\partial D}{\partial t}$, we can derive an expression for $\sigma$ as follows:
\begin{eqnarray}
    &&\frac{\partial D(\mathbf{r}(t))}{\partial t}|_{t=0}\\
\label{eq:lim1}
    &=& \lim_{\Delta t \rightarrow 0} \frac{d(\mathbf{r}(\Delta t),\mathbf{v})-d(\mathbf{r}(0),\mathbf{v})}{\Delta t}.
\end{eqnarray}
The first term of Equation \ref{eq:lim1} can be deformed as follows:
\begin{eqnarray}
    &&d(\mathbf{r}(\Delta t),\mathbf{v})\\
    &=& \int_{t_n}^{t_f}\mathrm{exp}(-\int_{t_n}^t \sigma(\mathbf{r}(s+\Delta t)ds)\sigma(\mathbf{r}(t+\Delta t)tdt\\
\label{eq:ddt1}
    &=& \int_{t_n}^{t_f}\mathrm{exp}(-\int_{t_n+\Delta t}^{t+\Delta t} \sigma(\mathbf{r}(s)ds)\sigma(\mathbf{r}(t+\Delta t)tdt.
\end{eqnarray}

We set $S(t_n,t):=\mathrm{exp}(-\int_{t_n}^t \sigma(\mathbf{r}(s))ds)\sigma(\mathbf{r}(t))dt$.
When $t_f$ takes a sufficiently large value until all of the light gets reflected, the Equations \ref{eq:ttf}, \ref{eq:stntf}, and \ref{eq:int_stntf} are valid.
\begin{eqnarray}
\label{eq:ttf}
    T(t_f)&=&0\\
\label{eq:stntf}
    S(t_n,t_f)&=&0\\
\label{eq:int_stntf}
    \int_{t_n}^{t_f}S(t_n,t_f)&=&1
\end{eqnarray}

With Equations \ref{eq:ttf}, \ref{eq:stntf}, and \ref{eq:int_stntf}, $d(\mathbf{r}(\Delta t),\mathbf{v})$ can be deformed as follows:

\begin{eqnarray}
    &&d(\mathbf{r}(\Delta t),\mathbf{v})\\
    &=& \int_{t_n}^{t_f}S(t_n+\Delta t,t+\Delta t)tdt\\
    &=& \int_{t_n}^{t_f}S(t_n+\Delta t,t+\Delta t)(t+\Delta t)dt -\int_{t_n}^{t_f}S(t_n+\Delta t,t+\Delta t)\Delta t dt\\
    &=& \int_{t_n+\Delta t}^{t_f+\Delta t}S(t_n+\Delta t,t)tdt -\Delta t\int_{t_n}^{t_f}S(t_n+\Delta t,t+\Delta t)dt\\
\label{eq:ddt2}
    &=& \int_{t_n}^{t_f}S(t_n+\Delta t,t)tdt + \int_{t_f}^{t_f+\Delta t}S(t_n+\Delta t,t)tdt \nonumber \\
    && \qquad\qquad\qquad\qquad\qquad\qquad\qquad -\int_{t_n}^{t_n+\Delta t}S(t_n+\Delta t,t)tdt - \Delta t.
\end{eqnarray}

We calculate the first term of Equation \ref{eq:ddt2} as follows:
\begin{eqnarray}
    &&\int_{t_n}^{t_f}S(t_n+\Delta t,t)tdt\\
    &=& \int_{t_n}^{t_f}\mathrm{exp}(-\int_{t_n+\Delta t}^t \sigma(\mathbf{r}(s))ds)\sigma(\mathbf{r}(t))tdt\\
    &=& \int_{t_n}^{t_f}T(t)T(t_n+\Delta t)\sigma(t_n+\mathbf{r}(t))tdt\\
    &=& T(t_n+\Delta t)d(\mathbf{r}(0), \mathbf{v}).
\end{eqnarray}

Using Equation \ref{eq:stntf}, the second term of Equation \ref{eq:ddt2} is equal to $0$.
The third term of Equation \ref{eq:ddt2} is calculated as follows:
\begin{eqnarray}
    &&\int_{t_n}^{t_n+\Delta t}S(t_n+\Delta t,t)tdt\\
\label{eq:deltatn}
    &=& T(t_n+\Delta t)\int_{t_n}^{t_n + \Delta t}T(t)\sigma(\mathbf{r}(t))tdt.
\end{eqnarray}

Same as in the main paper, we assume that $t_n$ is small enough to be valid at $T(t_n)=1$.
Equation \ref{eq:deltatn} converges to $\Delta t\sigma(\mathbf{r}(t_n))t_n $ as $\Delta t\rightarrow 0$.
Therefore, the Equation \ref{eq:lim1} can be deformed as follows:
\begin{eqnarray}
    &&\frac{\partial D(\mathbf{r}(t))}{\partial t}|_{t=0}\\
    &=& \lim_{\Delta t \rightarrow 0} \frac{d(\mathbf{r}(\Delta t),\mathbf{v})-d(\mathbf{r}(0),\mathbf{v})}{\Delta t}\\
    &=& \lim_{\Delta t \rightarrow 0}\frac{1}{\Delta t}\left[  T(t_n+\Delta t)d(\mathbf{r}(0),\mathbf{v}) - \Delta t\sigma(\mathbf{r}(t))t_n -\Delta t -d(\mathbf{r}(0),\mathbf{v}) \right]\\
    &=& \lim_{\Delta t \rightarrow 0}\left[  \frac{T(t_n+\Delta t)-1}{\Delta t}d(\mathbf{r}(0),\mathbf{v}) - \sigma(\mathbf{r}(t_n))t_n -1 \right]\\
    &=& \lim_{\Delta t \rightarrow 0}\left[  \frac{\mathrm{exp}(\Delta t \sigma(\mathbf{r}(t_n))-\mathrm{exp}(0)}{\Delta t}d(\mathbf{r}(0),\mathbf{v}) - \sigma(\mathbf{r}(t_n))t_n -1 \right]\\
    &=& \sigma(\mathbf{r}(t_n))d(\mathbf{r}(0),\mathbf{v}) - \sigma(\mathbf{r}(t_n))t_n -1\\
    &=& -1 + \left( D(\mathbf{r}(0)) - t_n \right)\sigma(\mathbf{r}(t_n)).
\end{eqnarray}

\section{Parameter selection for the shape of the auxiliary gradient}
\begin{figure}[ht]
    \begin{center}
     \includegraphics[width=0.5\linewidth]{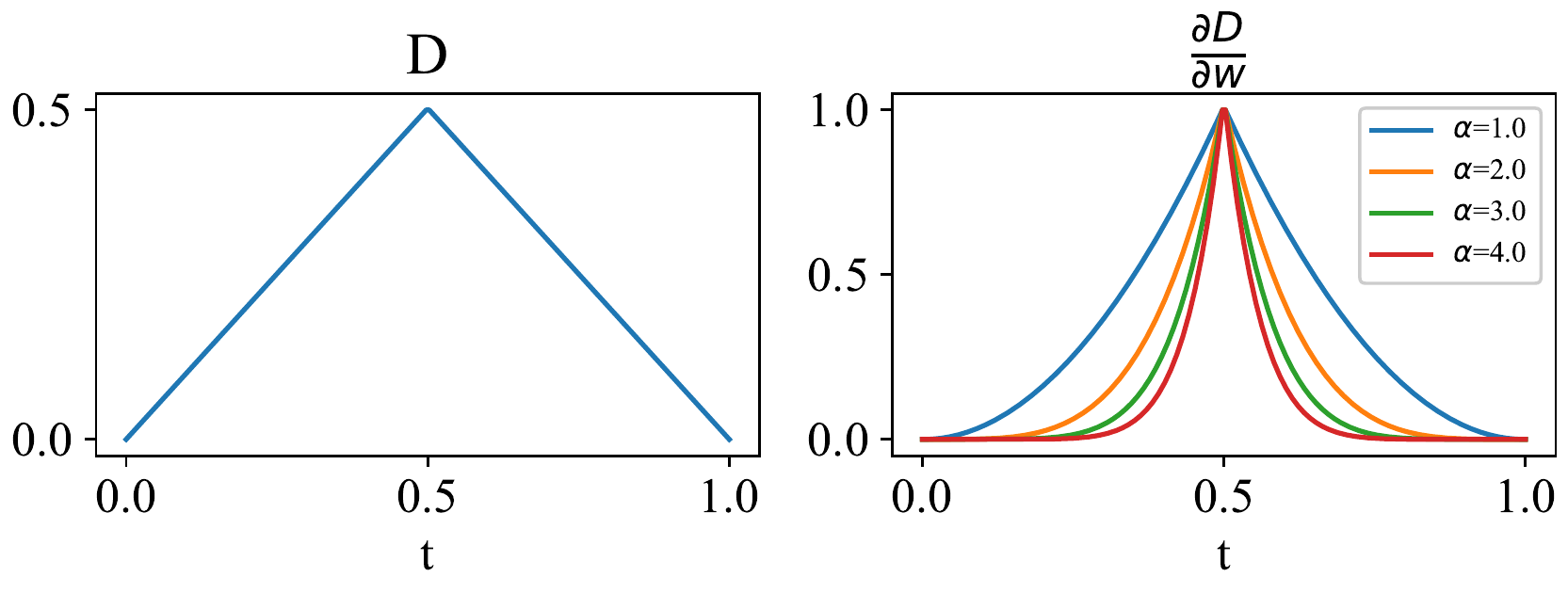}
     \end{center}
     \caption{Shape of auxiliary gradient for each $\alpha$}
     \label{fig:dDdwShape}
\end{figure}

Equation 15 in the main paper is a penalty term that constrains the shape of the auxiliary gradient by the hyperparameter $\alpha$.
Fig. \ref{fig:dDdwShape} shows that the auxiliary gradient becomes active in a narrower range than the distance field when $\alpha \leq 1$, and the shape becomes more concentrated near the cusps as $\alpha$ is larger.
This constraint leads to a unique set of auxiliary gradients.

\begin{figure}[ht]
    \begin{center}
     \includegraphics[width=0.8\linewidth]{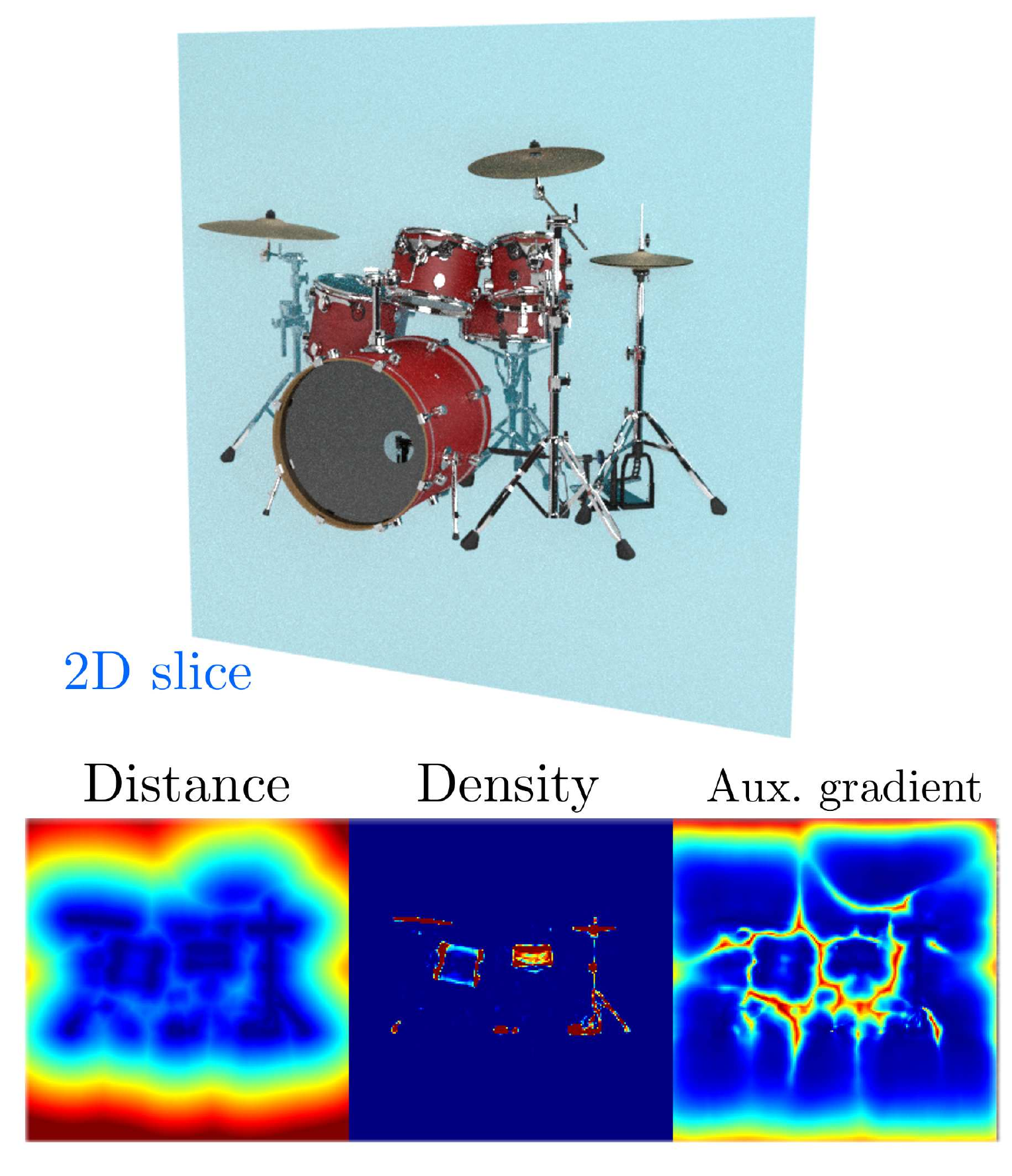}
     \end{center}
     \caption{Visualization of the 2D slice for distance, density, and auxiliary gradient.}
     \label{fig:auxgrad_field}
\end{figure}

\begin{figure}[ht]
    \begin{center}
     \includegraphics[width=0.8\linewidth]{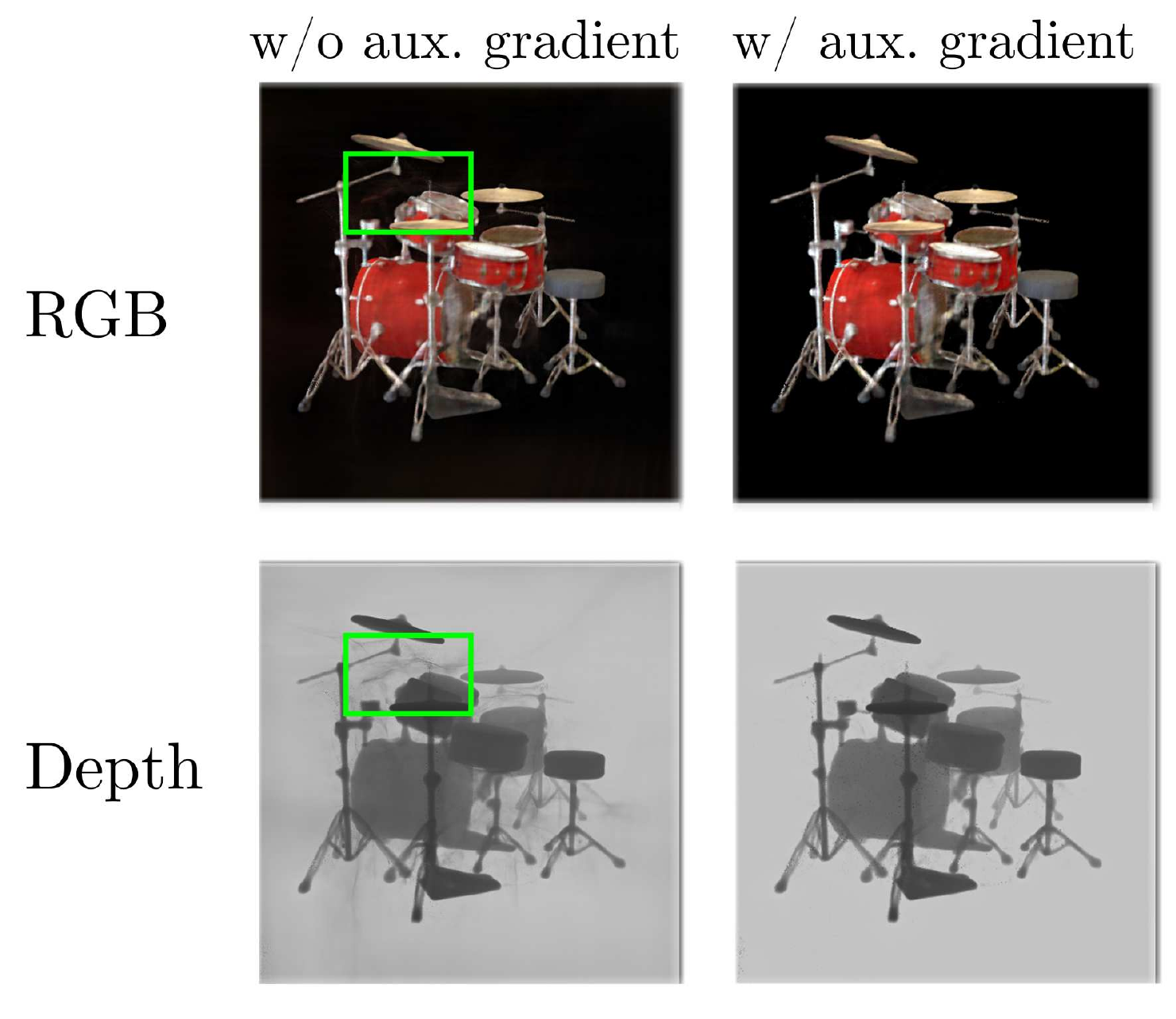}
     \end{center}
     \caption{The difference in rendering results with and without auxiliary gradients.}
     \label{fig:auxgrad_render}
\end{figure}

Fig. \ref{fig:auxgrad_field} is a colorized visualization of the distance field, density field, and auxiliary gradient in the 2D slice.
We can see that the auxiliary gradient becomes strongly activated near the cusp of the distance field, where the distances from several objects are similar.
Fig. \ref{fig:auxgrad_render} shows the difference in rendering results with and without auxiliary gradients.
Without auxiliary gradients, incorrect volume densities occur in the empty region.


\section{Evaluation of reconstruction performance in smoke scenes}
Most of the previous methods do not provide benchmarks for scenes with subjects containing smoke.
We produce a synthetic dataset for a smoke-subject scene where the density varies over a wide area, and we qualitatively evaluate the proposed method's performance.
As with the nerf synthetic dataset~\cite{NeRF}, the dataset consists of 100 viewpoints each in a hemispherical plane for train/valid data and 200 viewpoints in orbit for test data. For each shot, we record RGB and Transmittance information at a resolution of $800 \times 800$.

\begin{figure}[ht]
    \begin{center}
     \includegraphics[width=0.7\linewidth]{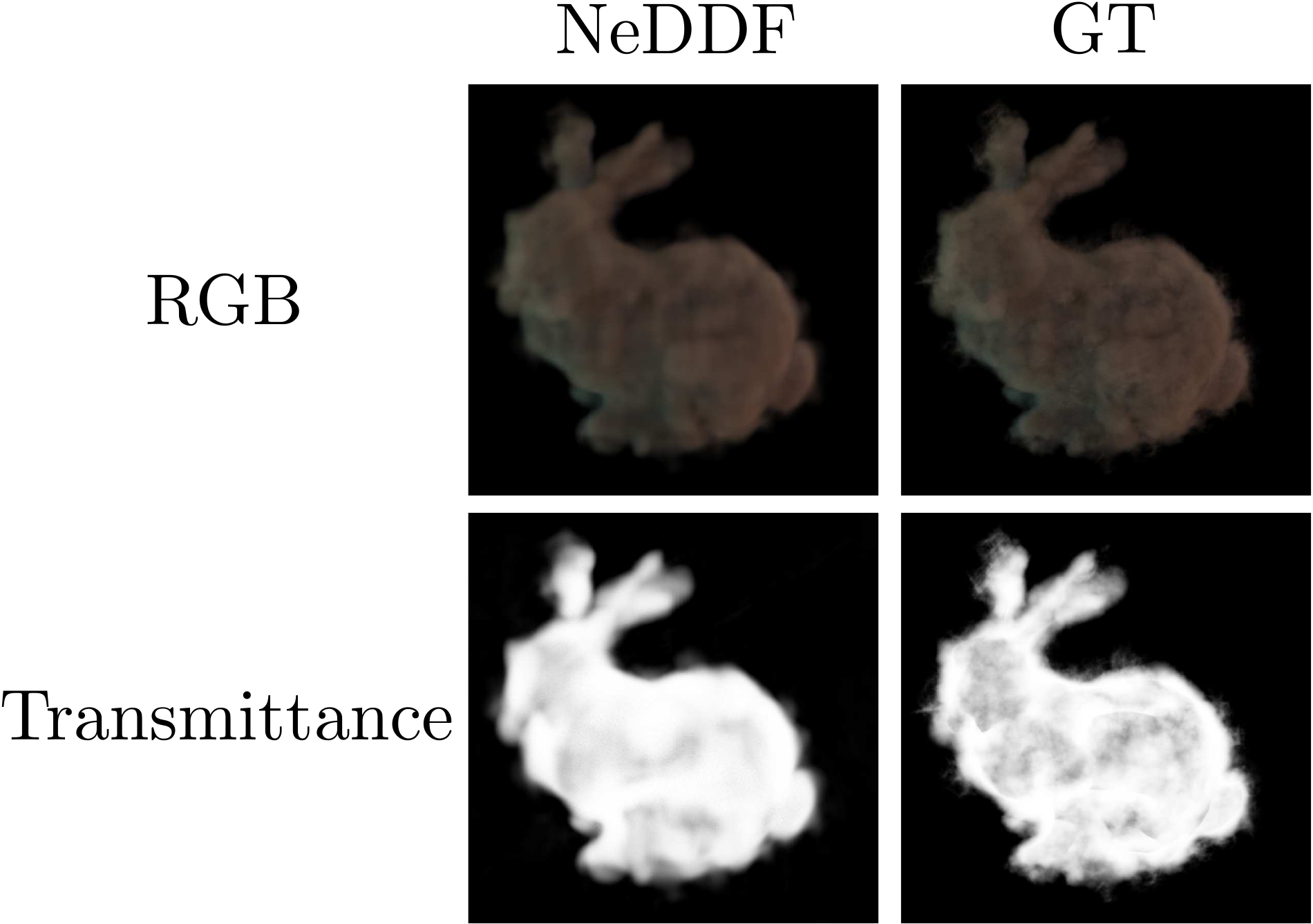}
     \end{center}
     \caption{Rendering results in smoke scene}
     \label{fig:result_smoke}
\end{figure}

\begin{figure}[ht]
    \begin{center}
     \includegraphics[width=0.7\linewidth]{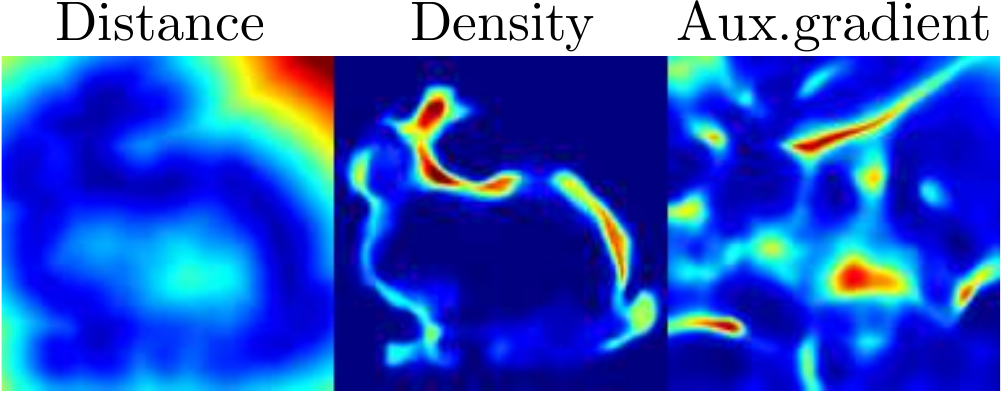}
     \end{center}
     \caption{Visualization of the 2D slice for distance, density, and auxiliary gradient in smoke scene.}
     \label{fig:result_field_smoke}
\end{figure}

Fig. \ref{fig:result_smoke} shows the rendered image from the test viewpoint. Our method achieves high-quality Novel View Synthesis even in smoke-like scenes.
Fig. \ref{fig:result_field_smoke} also shows the visualization results of the slices for the distance field, density field, and auxiliary gradient.
Since the distance field shows that the distance increases again inside the object, the field's estimation also holds true inside the object.
In addition, the minima of the distance field are larger than those in Fig. \ref{fig:auxgrad_field}, indicating that the assumption of expressing low density with large minima of the distance field works reasonably.
The density field actually expresses a translucent state rather than a bipolar one.


\section{Evaluation of localization performance in other scenes}
\begin{figure}[ht]
    \begin{center}
     \includegraphics[width=0.8\linewidth]{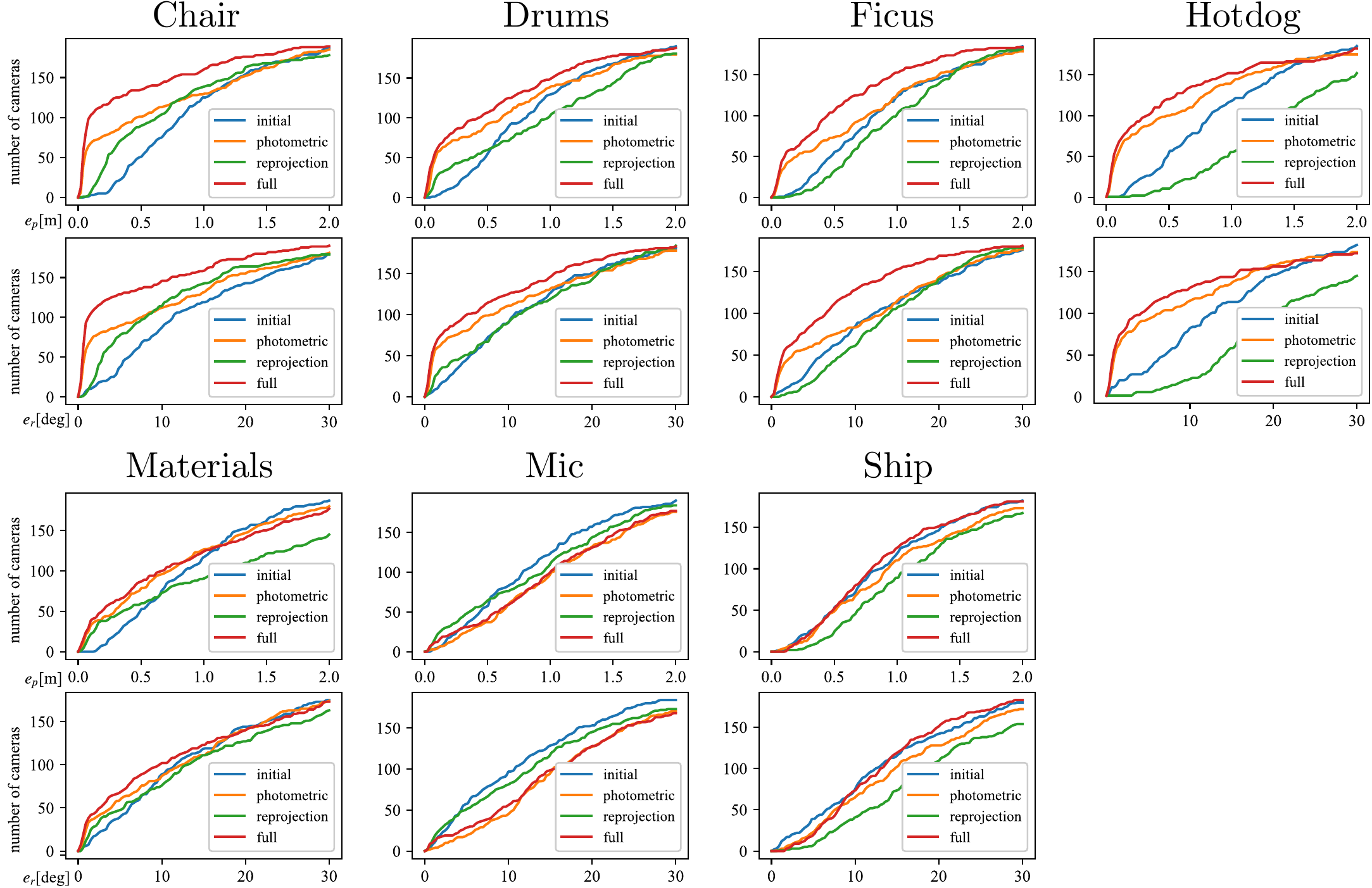}
     \end{center}
     \caption{Quantitative evaluation of camera poses estimation accuracy in other scenes. The horizontal axis represents the position and angle error, and the vertical axis represents the number of cameras recovered under the errors.}
     \label{fig:err_count_full}
\end{figure}

For other scenes in the NeRF synthetic dataset~\cite{NeRF}, we verify the camera localization performance in the same way as in experiment (b) of main paper.
Fig. \ref{fig:err_count_full} plots the number of camera postures for which the position and angular errors are lower than the threshold values for each scene.
In all cases, the use of the reprojection error improves performance more than the use of the photometric error alone, as in iNeRF\cite{iNeRF}.
Even in cases such as Drums and Ficus scenes, where the optimization result with only the reprojection error is worse than the initial value, we can see an improvement of performance due to the increase in the common area of the field of view.
On the other hand, in scenes where the uniqueness of color information is not sufficient, such as the Materials scene, reprojection error may degrade performance due to mismatches between corresponding points.
In the case where many local solutions for photometric errors exist, such as Mic and Ship scenes, the use of reprojection error does not avoid the local solutions and does not improve the performance.
We believe that we can improve such scenes by propagating unique features other than color to the empty regions in the same way as color fields.

\end{document}